\documentclass[runningheads]{llncs}
\usepackage{graphicx}
\usepackage{amsmath,amssymb} 
\usepackage{color}
\usepackage[width=122mm,left=12mm,paperwidth=146mm,height=193mm,top=12mm,paperheight=217mm]{geometry}

\usepackage{subcaption}
\usepackage{epstopdf}

\begin{document}
\pagestyle{headings}
\mainmatter
\def\ECCV16SubNumber{}  

\title{Deep Disentangled Representations for Volumetric Reconstruction} 

\author{Edward Grant\inst{1}, Pushmeet Kohli\inst{2}, Marcel van Gerven\inst{3}}


\institute{1   Radboud University   \email{edward339@gmail.com}\\
2   Microsoft Research   \email{pkohli@microsoft.com}\\
3   Radboud University   \email{m.vangerven@donders.ru.nl}\\
}

\maketitle

\begin{abstract}

We introduce a convolutional neural network for inferring a compact disentangled graphical description of objects from 2D images that can be used for volumetric reconstruction. The network comprises an encoder and a twin-tailed decoder. The encoder generates a disentangled \textit{graphics code}. The first decoder generates a volume, and the second decoder reconstructs the input image using a novel training regime that allows the \textit{graphics code} to learn a separate representation of the 3D object and a description of its lighting and pose conditions. We demonstrate this method by generating volumes and disentangled graphical descriptions from images and videos of faces and chairs. 
\end{abstract}

\section{Introduction}
Images depicting natural objects are 2D representations of an underlying 3D structure from a specific viewpoint in specific lighting conditions.

This work demonstrates a method for recovering the underlying 3D geometry of an object depicted in a single 2D image or video. To accomplish this we first encode the image as a separate description of the shape and transformation properties of the input such as lighting and pose. The shape description is used to generate a volumetric representation that is interpretable by modern rendering software. 

State of the art computer vision models perform recognition by learning hierarchical layers of feature detectors across overlapping sub-regions of the input space. Invariance to small transformations to the input is created by sub-sampling the image at various stages in the hierarchy. 

In contrast, computer graphics models represent visual entities in a canonical form that is disentangled with respect to various realistic transformations in 3D, such as pose, scale and lighting conditions. 2D images can be rendered from the graphics code with the desired transformation properties. 

A long standing hypothesis in computer vision is that vision is better accomplished by inferring such a disentangled graphical representation from 2D images. This process is known as `de-rendering' and the field is known as `vision as inverse graphics'~\cite{yuille2006vision}.

One obstacle to realising this aim is that the de-rendering problem is ill-posed. The same 2D image can be rendered from a variety of 3D objects. This uncertainty means that there is normally no analytical solution to de-rendering. There are however, solutions that are more or less likely, given an object class or the class of all natural objects. 

Recent work in the field of vision as inverse graphics has produced a number of convolutional neural network models that accomplish de-rendering~\cite{kulkarni2015deep,tatarchenko2015single,yang2015weakly}. Typically these models follow an encoding / decoding architecture. The encoder predicts a compact 3D graphical representation of the input. A control signal is applied corresponding with a known transformation to the input and a decoder renders the transformed image. We use a similar architecture. However, rather than rendering an image from the graphics code, we generate a full volumetric representation.

Unlike the disentangled graphics code generated by existing models, which is only renderable using a custom trained decoder, the volumetric representation generated by our model is easily converted to a polygon mesh or other professional quality 3D graphical format. This allows the object to be rendered at any scale and with other rendering techniques available in modern rendering software. 

\section{Related work}
Several models have been developed that generate an disentangled representation given a 2D input, and output a new image subject to a transformation. 

Kulkarni \textit{et al}. proposed the Deep Convolutional Inverse Graphics Network (DC-IGN) trained using Stochastic Gradient Variational Bayes~\cite{kulkarni2015deep}. This model encodes a factored latent representation of the input that is disentangled with respect to changes in azimuth, elevation and light source. A decoder renders the graphics code subject to the desired transformation as a 2D image. Training is performed with batches in which only a single transformation or the shape of the object are different. The activations of the graphics code layer chosen to represent the static parameters are clamped as the mean of the activations for that batch on the forward pass. On the backward pass the gradients for the corresponding nodes are set to their difference from this mean. The method is demonstrated by generating chairs and face images transformed with respect to azimuth, elevation and light source.

Tatarchenko \textit{et al}. proposed a similar model that is trained in a fully supervised manner~\cite{tatarchenko2015single}. The encoder takes a 2D image as input and generates a graphics code representing a canonical 3D object form. A signal is added to the code corresponding with a known transformation in 3D and the decoder renders a new image corresponding with that transformation. This method is also demonstrated by generating rotated images of cars and chairs. 

Yang \textit{et al}. demonstrated an encoder / decoder model similar to the above but utilize a recurrent structure to account for long-term dependencies in a sequence of transformations, allowing for realistic re-rendering of real face images from different azimuth angles~\cite{yang2015weakly}. 

Spatial Transformer Networks (STN) allow for the spatial manipulation of images and data within a convolutional neural network~\cite{jaderberg2015spatial}. The STN first generates a transformation matrix given an input, creates a grid of sampling points based on the transformation and outputs samples from the grid. The module is trained using back-propagation and transforms the input with an input dependent affine transformation. Since the output sample can be of arbitrary size, these modules have been used as an efficient down-sampling method in classification networks. STNs transform existing data by sampling but they are not generative, so cannot make predictions about occluded data, which is necessary when predicting 3D structure. 

Girdhar \textit{et al}. and Rezende \textit{et al}.  present methods for volumetric reconstructing from 2D images but do not generate disentangled representations~\cite{girdhar2016learning,rezende2016unsupervised}. 

The contribution of this work is an encoding / decoding model that generates a compact graphics code from 2D images and videos that is disentangled with respect to shape and the transformation parameters of the input, and that can also be used for volumetric reconstruction. To our knowledge this is the first work that generates a disentanlged graphical representation that can be used to reconstruct volumes from 2D images. In addition, we show that Spatial Transformer Networks can be used to replace max-pooling in the encoder as an efficient sampling method.  We demonstrate this approach by generating a compact disentangled graphical representation from single 2D images and videos of faces and chairs in a variety of viewpoint and lighting conditions. This code is used to generate volumetric representations which are rendered from a variety of viewpoints to show their 3D structure.

\section{Model}
\subsection{Architecture}
\begin{figure}[!t]
\centering
\includegraphics[width=\textwidth]{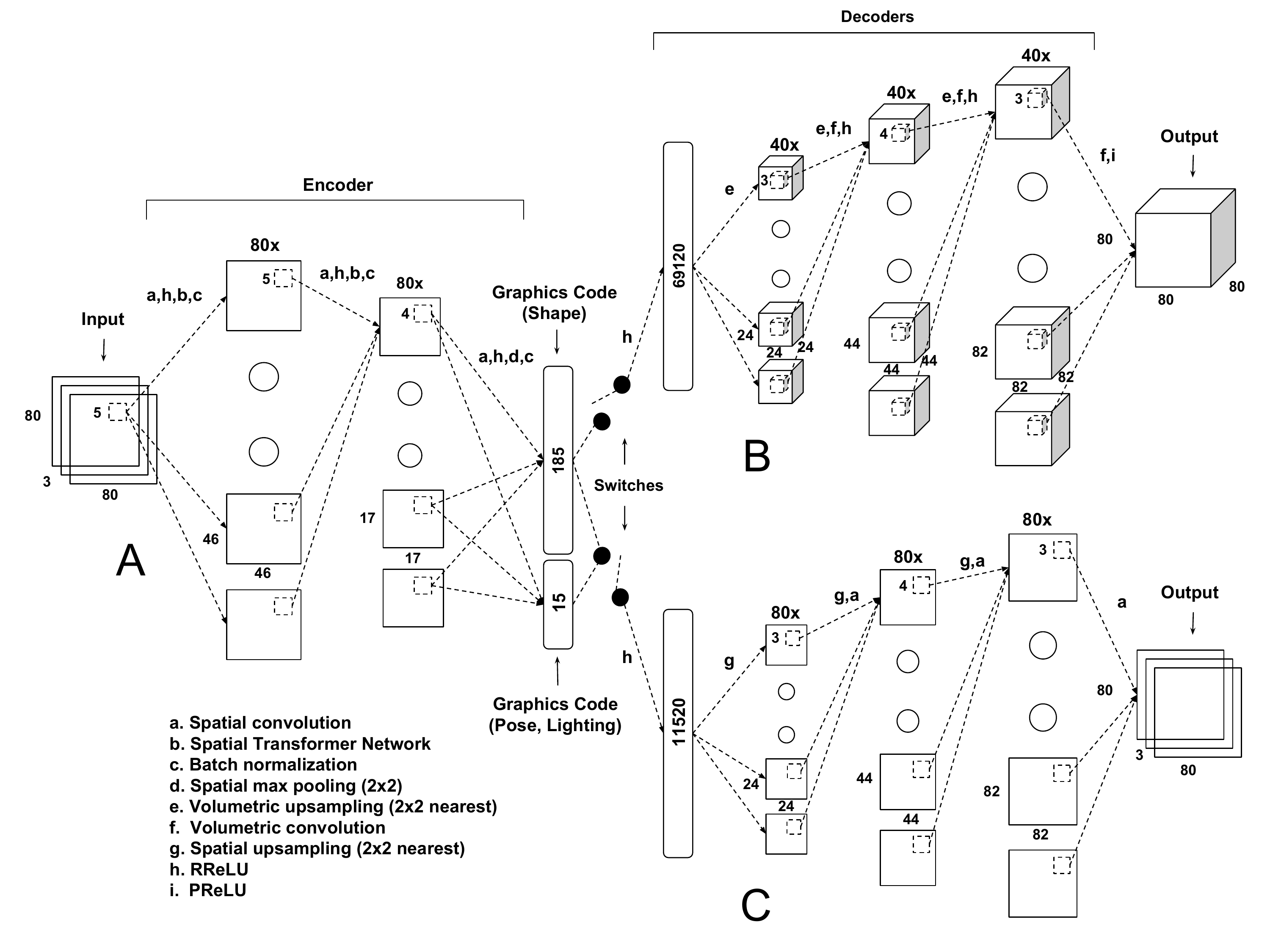}
\caption{\textbf{Network architecture:} The network consists of an encoder (A), a volume decoder (B) and an image decoder (C). The encoder takes as input a 2D image and generates a 3D \textit{graphics code} through a series of spatial convolutions, down-sampling Spatial Transformer Networks and max pooling layers. This code is split into a \textit{shape code} and a \textit{transformation code}. The volume decoder takes the \textit{shape code} as input and generates a prediction of the volumetric contents of the input. The image decoder takes the \textit{shape code} and the \textit{transformation code} as input and reconstructs the input image.}
\label{fig:network}
\end{figure}

As shown in Figure~\ref{fig:network}, the network has one encoder, a \textit{graphics code} layer and two decoders. The \textit{graphics code} layer is separated into a \textit{shape code} and a \textit{transformation code}. The encoder takes as input an 80 $\times$ 80 pixel color image and generates the \textit{graphics code} following a series of convolutions, point-wise randomized rectified linear units (RReLU)~\cite{xu2015empirical}, down-sampling Spatial Transformer Networks and max pooling. Batch normalization layers are used after each convolutional layer to speed up training and avoid problems with exploding and vanishing gradients~\cite{ioffe2015batch}. 

The two decoders are connected to the \textit{graphics code} by switches so that the message from the \textit{graphics code} is passed to either one of the decoders. The first decoder is the volume decoder. The volume decoder takes the \textit{shape code} as input and generates an $80 \times 80 \times 80$ voxel volumetric prediction of the encoded shape. This is accomplished by a series of volumetric convolutions, point-wise RReLU and volumetric up-sampling. A parametric rectified linear unit (PReLU)~\cite{he2015delving} is substituted for the RReLU in the output layer. This is done to avoid the saturation problems with rectified linear units early in training but allows for learning an activation threshold later in training, corresponding with the positive-valued output targets.

\begin{figure}
\centering
\includegraphics[width=0.4\textwidth]{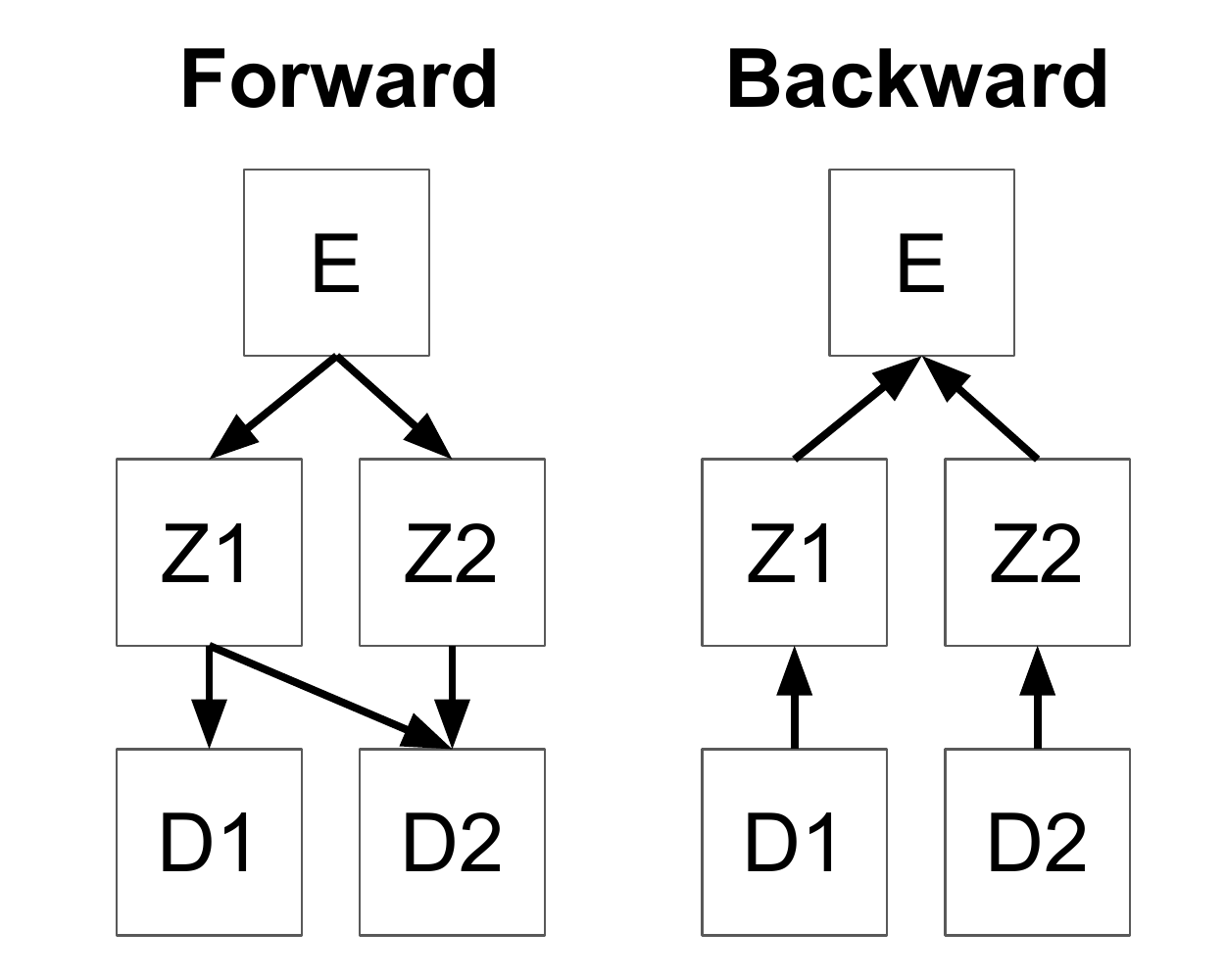}
\caption{\textbf{Network training:} In the forward pass the \textit{shape code} (Z1) and the \textit{transformation code} (Z2) receive a signal from the encoder (E). The volume decoder (D1) receives input only from the \textit{shape code}. The image decoder (D2) receives input from the \textit{shape code} and the \textit{transformation code}. On the backward pass the signal from the image decoder to the \textit{shape code} is suppressed to force it to only represent shape.}
\label{fig:training}
\end{figure}

The second decoder reconstructs the input image with the correct pose and lighting, showing that pose and lighting parameters of the input are contained in the \textit{graphics code}. The image decoder takes as input both the \textit{shape code} and the \textit{transformation code}, and generates a reconstruction of the original input image. This is accomplished by a series of spatial convolutions, point-wise RReLU, spatial up-sampling and point-wise PReLU in the final layer. During training, the backward pass from the image decoder to the \textit{shape code} is blocked (see Figure~\ref{fig:training}). This encourages the \textit{shape code} to only represent shape, as it only receives an error signal from the volume decoder.

The volume decoder only requires knowledge about the shape of the input since it generates binary volumes that are invariant to pose and lighting. However, the image decoder must generate a reconstruction of the original image which is not invariant to shape, pose or lighting. Both decoders have access to the \textit{shape code} but only the image decoder has access to the \textit{transformation code}. This encourages the network to learn a \textit{graphics code} that is disentangled with respect to shape and transformations. 

The network can be trained differently depending on whether pose and lighting conditions need to be encoded. If the only objective is to generate volumes from the input then the image decoder can be switched off during training. In this case the \textit{graphics code} will learn to be invariant to viewpoint and lighting. If the volume decoder and image decoder are both used during training the \textit{graphics code} learns a disentangled representation of shape and transformations.

\subsection{Spatial transformer networks}

Spatial Transformer Networks (STNs) perform input dependent geometric transformations on images or sets of feature maps~\cite{jaderberg2015spatial}. There are two STNs in our model (see Figure~\ref{fig:network}).

Each STN comprises a localisation network, a grid generator and sampling grid. The localisation network takes the activations of the previous layer as input and regresses the parameters of an affine transformation matrix. The grid generator generates a sampling grid of ($x,y$) coordinates corresponding with the desired height and width of the output. The sampling grid is obtained by multiplying the generated grid with the transformation matrix. In our model this takes the form:
\begin{align}
\begin{pmatrix}
  x_{i}^{s} \\
  y_{i}^{s} 
 \end{pmatrix}
= \mathcal{T}_{\theta} (G_{i})=
\begin{bmatrix}
  \theta_{11} & \theta_{12} & \theta_{13} \\
  \theta_{21}& \theta_{22} & \theta_{23}
 \end{bmatrix}
\begin{pmatrix}
  x_{i}^{t} \\
  y_{i}^{t} \\
  1
 \end{pmatrix}
\end{align}

\noindent
Where ($x_{i}^{t},y_{i}^{t}$) are the generated grid coordinates and ($x_{i}^{s},y_{i}^{s}$) define the sample points. The transformation matrix $\mathcal{T}_{\theta}$ allows for cropping, scale, translation, scale, rotation and skew. Cropping and scale, in particular allow the STN to focus on the most important region in a feature map. 

STNs have been shown to improve performance in convolutional network classifiers by modelling attention and transforming feature maps. Our model uses STNs in a generative setting to perform efficient down-sampling and assist the network in learning invariance to pose and lighting. 

The first STN in our model is positioned after the first convolutional layer. It uses a convolutional neural network to regress the transformation coefficients. This localisation network consists of four $5\times5$ convolutional layers, each followed by batch normalization and the first three also followed by $2\times2$ max pooling. 

The second STN in our model is positioned after the second convolutional layer and regresses the transformation parameters with a convolutional network consisting of two $5\times5$ an one $6\times6$ convolutional layers each followed by batch normalization and the last two also by $2\times2$ max pooling.

\subsection{Data}
The model was trained using $16,000$ image-volume pairs generated from the Basel Face Model \cite{paysan20093d}. Images of size $80\times80$ were rendered in RGB from five different azimuth angles and three ambient lighting settings. Volumes of size $80\times80\times80$ were created by discretizing the triangular mesh generated by the Basel Face Model. 
\section{Experimental Results}

\subsection{Training}
We evaluated the model's volume prediction capacity by training it on $16,000$ image-volume pairs. Each example pair was shown to the network only once to discourage memorization of the training data. 

Training was performed using the Torch framework on a single NVIDIA Tesla K80 GPU. Batches of size 10 were given as input to the encoder and forward propagated through the network. The mean-squared error of the predicted and target volumes was calculated and back-propagated using the Adam learning algorithm \cite{kingma2014adam}. The initial learning rate was set to $0.001$.

\subsection{Volume Predictions from Images of Faces}
In this experiment we used the network to generate volumes from a single 2D images. The network was presented with unseen face images as input and generated 3D volume predictions. The image decoder was not used in this experiment.

The predicted volumes were binarized with a threshold of $0.01$. A triangular mesh was generated from the coordinates of active voxels using Delaunay triangulation. The patch was smoothed and the resulting image rendered using OpenGL and Matlab's {\tt trimesh} function. 

\begin{figure}[!t]
  \hspace{-8mm}
  \begin{tabular}[b]{cc}
            \begin{subfigure}[b]{1.1\columnwidth}
      \includegraphics[width=\textwidth]{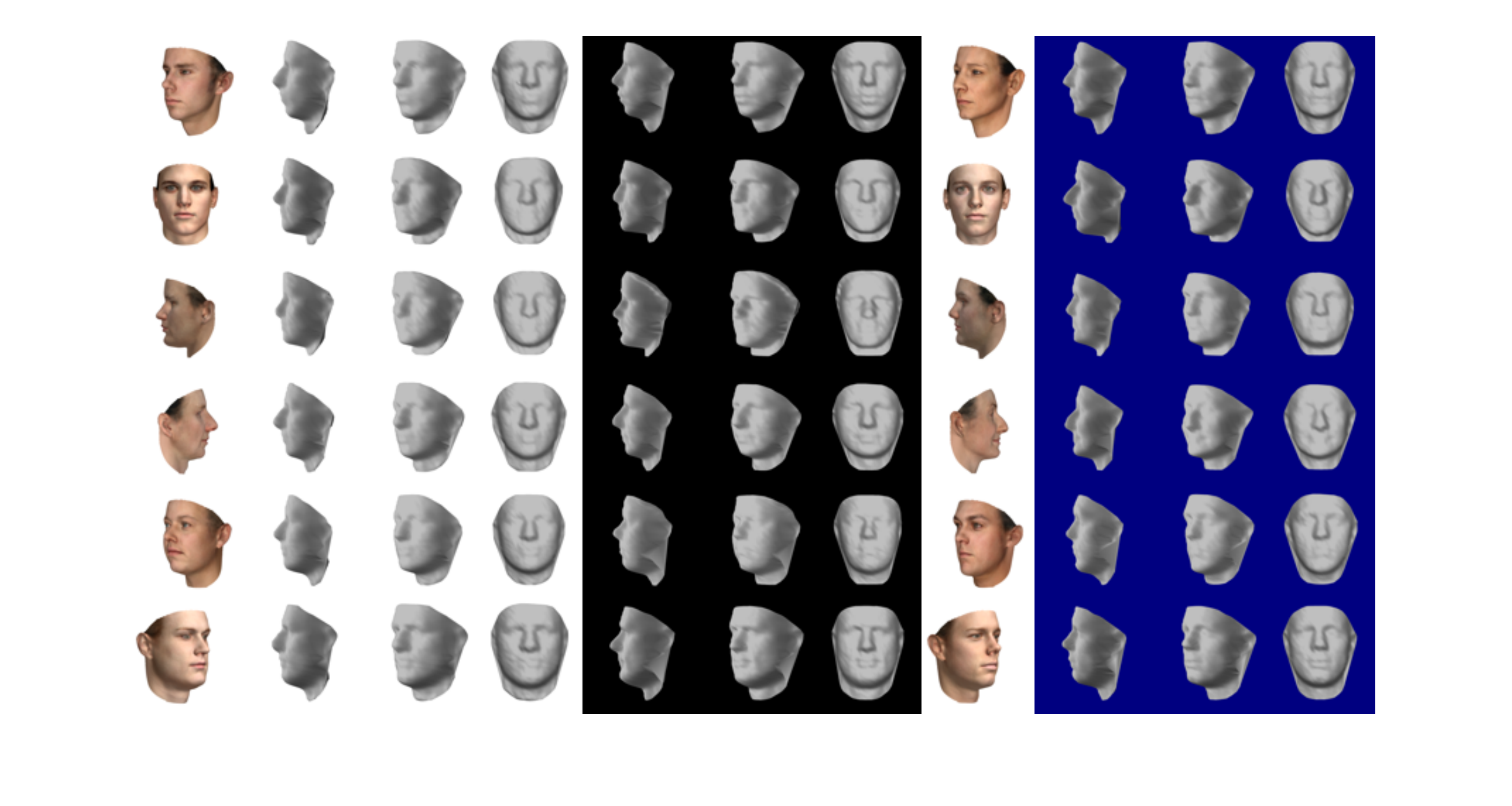}
      \vspace*{-10mm}
      \caption{}
      \label{fig:A}
    \end{subfigure}\\
      \hspace{-5mm}
      \begin{subfigure}[b]{0.75\columnwidth}
        \includegraphics[width=1\textwidth]{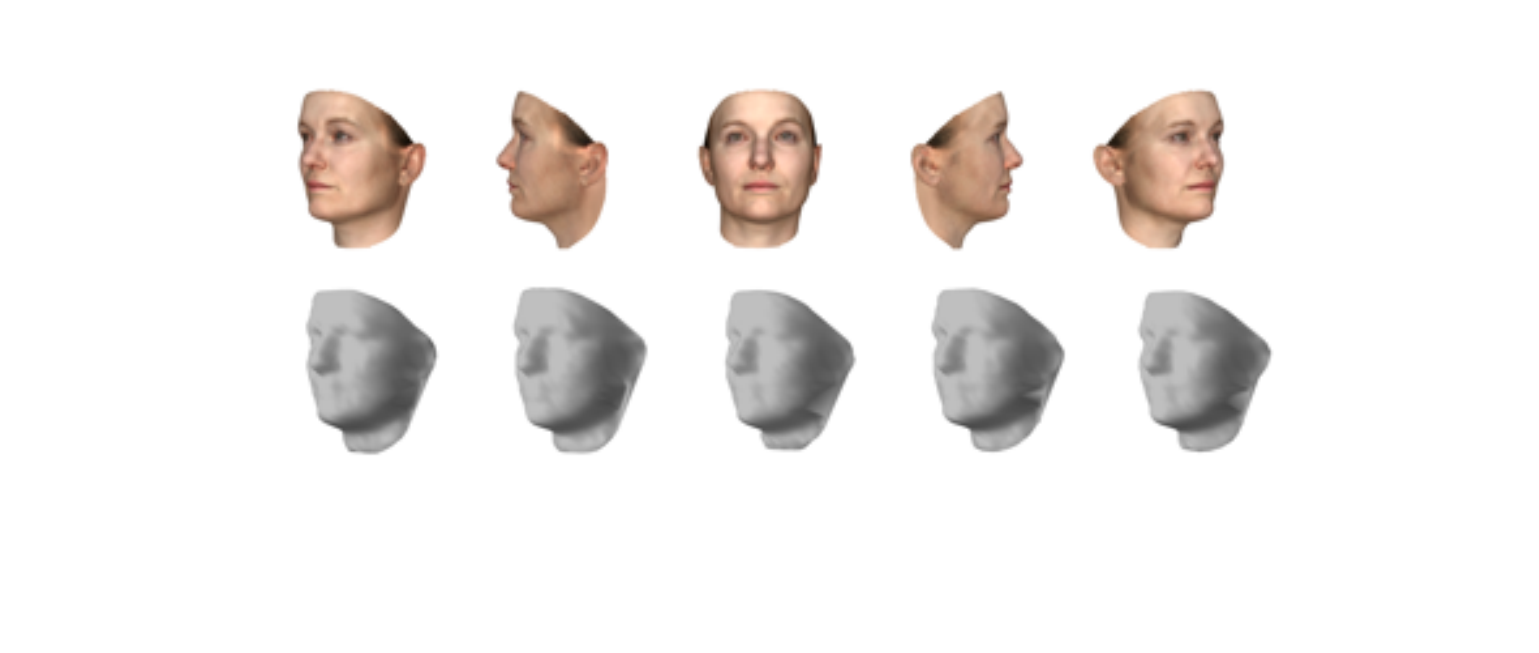}
        \vspace*{-10mm}
        \caption{}
        \label{fig:B}
                 \vspace*{-1.1mm}
      \end{subfigure}
          \hspace{-17mm}
        \begin{subfigure}[b]{0.52\columnwidth}
        \includegraphics[width=\textwidth]{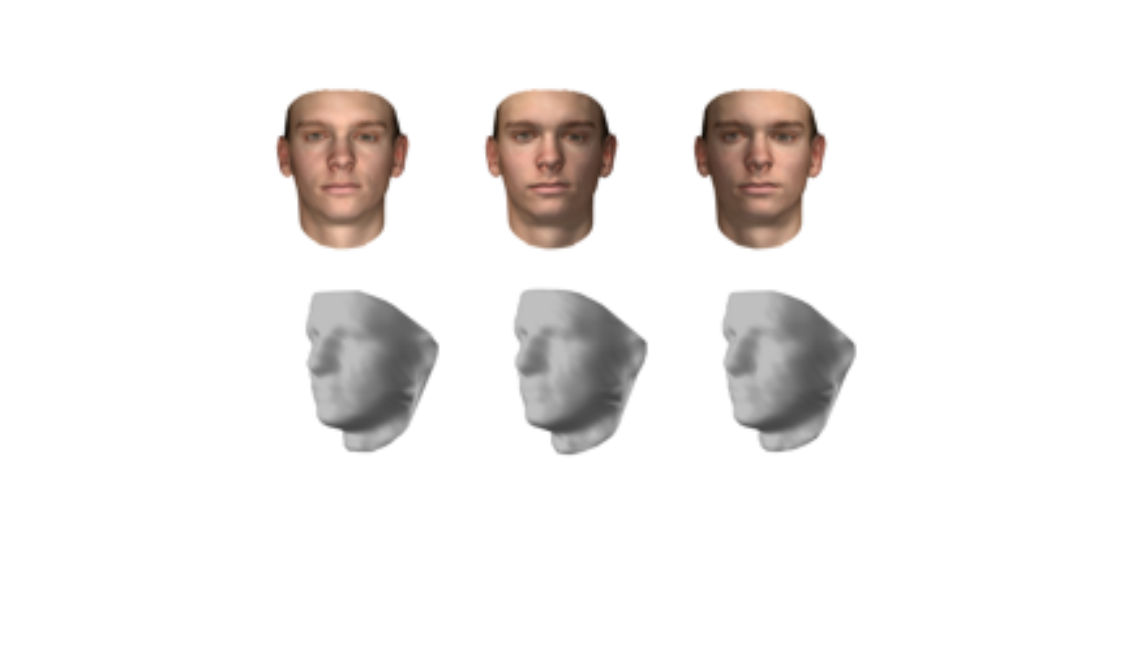}
        \vspace*{-10mm}
        \caption{}
        \label{fig:C}
      \end{subfigure}
      \end{tabular}
  \caption{\textbf{Generated volumes:} Qualitative results showing the volume predicting capacity of the network on unseen data. (a) First column: network inputs. Columns 2-4 (white): network predictions shown from three viewpoints. Columns 5-7 (black): ground truth from the same viewpoints. Column 8: nearest neighbour image. Columns 9-11 (blue): nearest neighbour image ground truth. (b) Each column is an input/output pair. The inputs are in the first row. Each input is the same face viewed from a different position. The generated volumes in the second row are shown from the same viewpoint for comparison. (c) Each column is an input/output pair. The inputs are in the first row. Each input is the same face in different lighting conditions.} 
    \label{fig:ABC}
\end{figure}

Figure~\ref{fig:ABC}(a) shows the input image, network predictions, ground truth, nearest neighour in the input space and the ground truth of the nearest neighour. The nearest neighbour was determined by searching the training images for the image with the smallest pixel-wise distance to the input. The generated volumes are visibly different depending on the shape of the input.

Figure~\ref{fig:ABC}(b) shows the network output for the same input presented from different viewpoints. The images in the first row are the inputs to the network and the second row contains the volumes generated from each input. These are shown from the same viewpoint for comparison. The generated volumes are visually very similar, showing that the network generated volumes that are invariant to the pose of the input.

Figure~\ref{fig:ABC}(c) shows the network output for the same face presented in different lighting conditions. The first row images are the inputs and the second row are the generated volumes also shown from the same viewpoint for comparison. These volumes are also visually very similar to each other showing that the network output appears invariant to lighting conditions in the input.

\subsection{Nearest Neighbour Comparison}
The network's quantitative performance was benchmarked using a nearest neighbour test. A test set of 200 image / volume pairs was generated using the Basel Face Model (ground truth). The nearest neighbour to each test image in the training set was identified by searching for the training set image with the smallest pixel-wise Euclidean distance to the test set image (nearest neighbour). The network generated a volume for each test set input (prediction).

Nearest neighbour error was determined by measuring the mean voxel-wise Euclidean distance between the ground truth and nearest neighbour volumes. Prediction error was determined by measuring the mean voxel-wise Euclidean distance between the ground truth volumes and the predicted volumes. 

A paired-samples t-test was conducted to compare error score in predicted and nearest neighbour volumes. There was a significant difference in the error score for predictions ($M=0.0096$, $SD=0.0013$) and nearest neighbours ($M=0.017$, $SD=0.0038$) conditions; $t(199)=-21.5945$,$ p=4.7022e-54$. 

These results show that network is better at predicting volumes than using the nearest neighbour.

\subsection{Internal Representations}
In this experiment we tested the ability of the encoder to generate a \textit{graphics code} that can be used to generate a volume that is invariant to pose and lighting. Since the volume encoder doesn't need pose and lighting information we didn't use the image decoder in this experiment.

To test the invariance of the encoder with respect to pose, lighting and shape we re-trained the model without using batch normalization. Three sets of 100 image batches were prepared where two of these parameters were clamped and the target parameter was different. This makes it possible to measure the variance of activations for changes in pose, lighting and shape. The set-wise mean of the mean variance of activations in each batch was compared for all layers in the network.

\begin{figure}[!t]
\hspace{-4mm}
  \begin{tabular}[b]{cc}
    \begin{subfigure}[b]{0.6\columnwidth}
      \includegraphics[width=\textwidth]{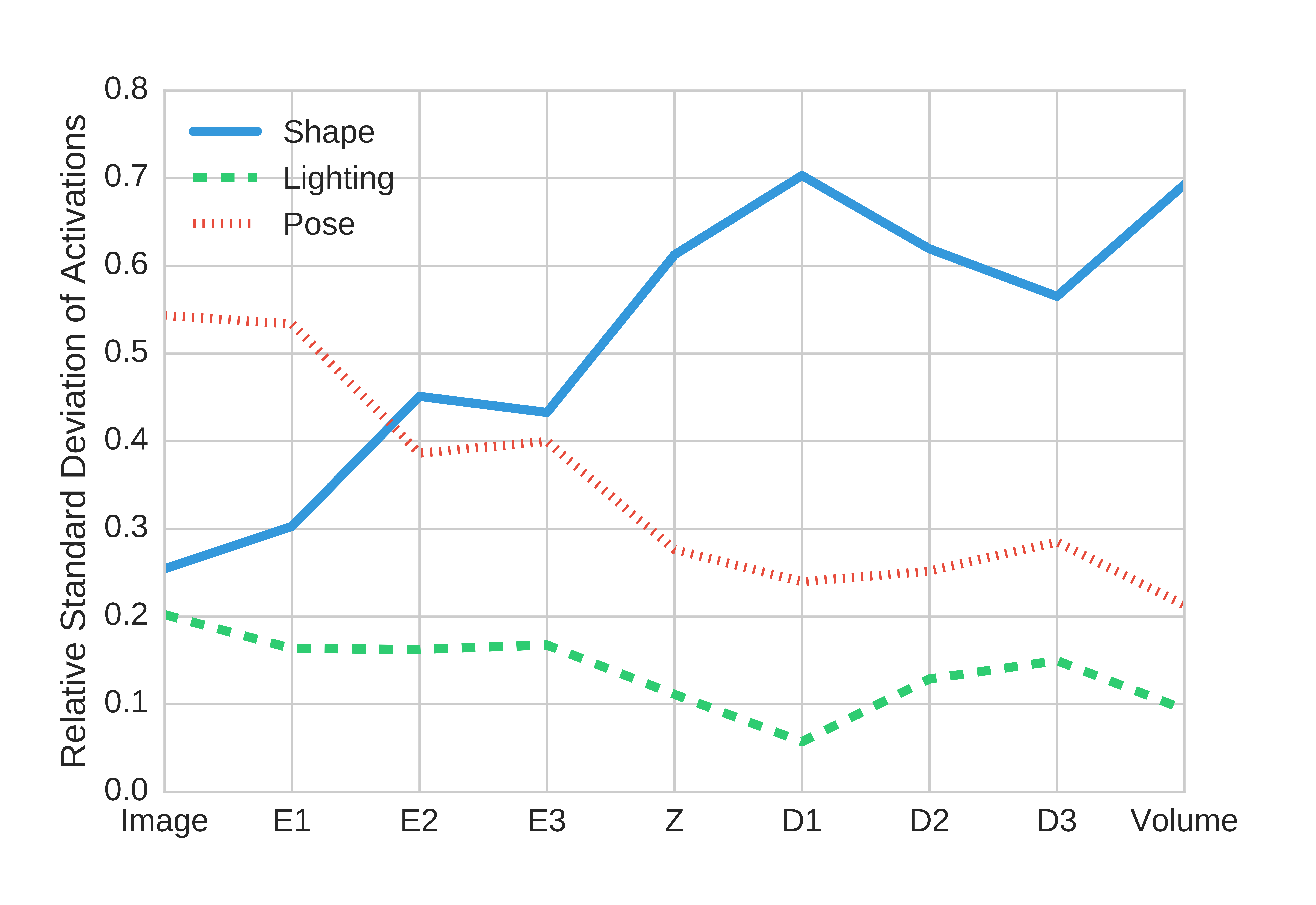}
 \vspace*{-8mm}
      \caption{}
      \label{fig:A}
    \end{subfigure}
\hspace{-6mm}
    \begin{tabular}[b]{c}
      \begin{subfigure}[b]{0.5\columnwidth}
        \includegraphics[width=\textwidth]{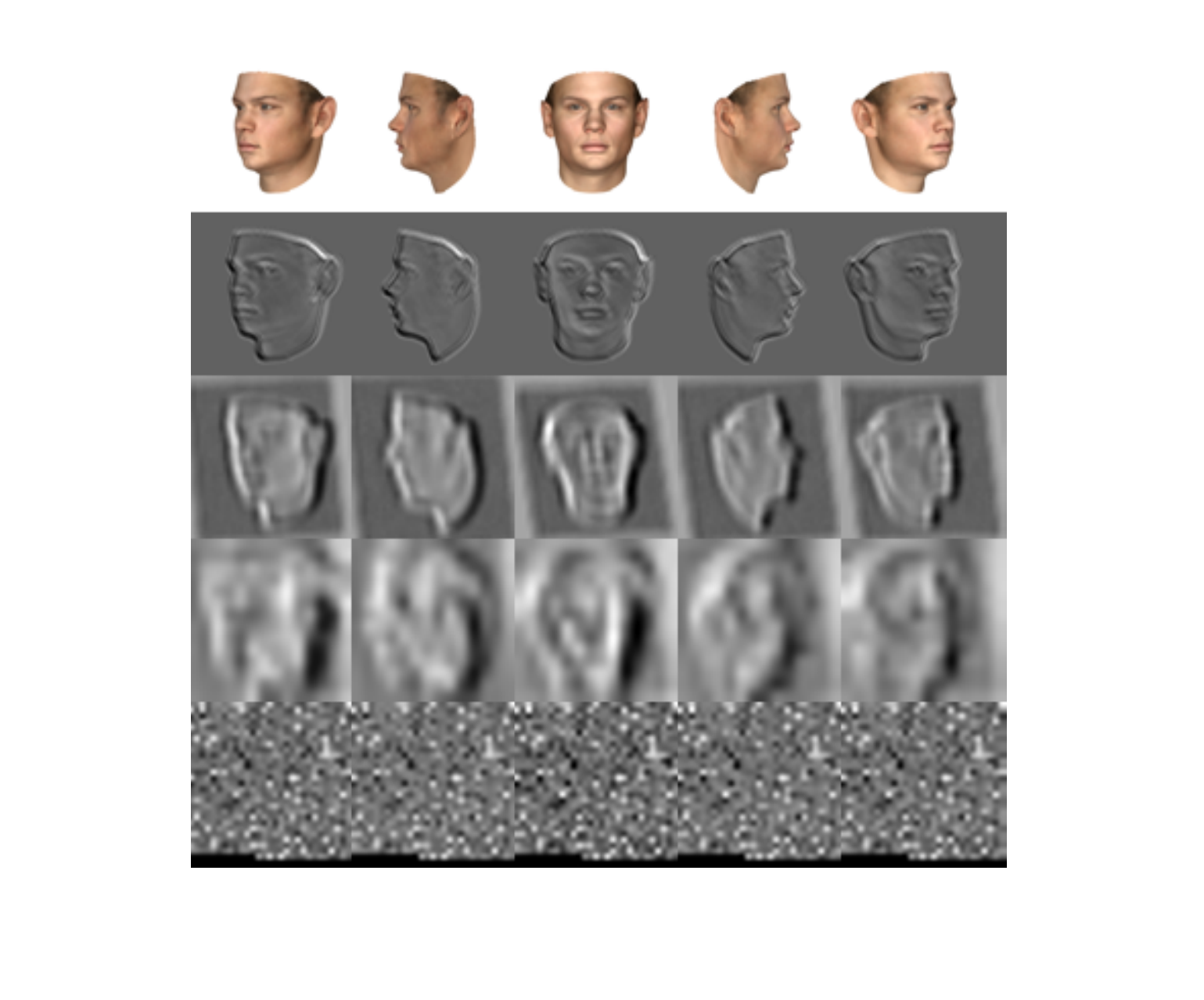}
        \vspace*{-10mm}
        \caption{}
        \label{fig:B}
      \end{subfigure}\\
    \end{tabular}
    \hspace{-10mm}
      \end{tabular}
    \hspace{-10mm}
\caption{\textbf{Invariance to pose and lighting:} (a) The relative mean standard deviation (SD) of activations in each network layer is compared for changes in shape, pose and lighting. Image is the input image, E1-E3 are the convolutional encoder layers, Z is the \textit{graphics code}, D1-D3 are the convolutional decoder layers and Volume is the generated volume. In the input, changes to pose account for the highest SD. By the second convolutional layer the network is more sensitive to changes in shape than pose or lighting. The \textit{graphics code} is much more sensitive to shape than pose or lighting. (b) The first row is five images of the same face from different viewpoints. Rows 2-4 show sampled encoder activations for the input image at the top of each column. The last row shows sampled \textit{graphics code} activations reshaped into a square.}
    \label{fig:invariance}
\end{figure}

Figure~\ref{fig:invariance}(a) shows that the network's heightened sensitivity to shape relative to pose and lighting begins in the second convolutional layer. There is a sharp increase in sensitivity to shape in the \textit{graphics code}, which is much more sensitive to shape than pose or lighting, and more sensitive to pose than lighting. This relative invariance to pose and lighting is retained in the volume decoder.

Figure~\ref{fig:invariance}(b) shows a visual representation of the activations for the same face with different poses. The effect of the first STN can be seen in the second convolutional layer activations which are visibly warped. The difference in the warp depending on the pose of the face suggests that the STNs may be helping to create invariance to pose later in the network. The example input images have a light source which is directed from the left of the camera. The second convolutional layer activations show a dark area on the right side of each face which is less evident in the first convolutional layer, suggesting that shadowing is an important feature for predicting the 3D shape of the face. 

\subsection{Disentangled Representations}
In this experiment we tested the network's ability to generate a compact 3D description of the input that is disentangled with respect to the shape of the object and transformations such as pose and lighting. 

In order to generate this description we used the same network as in the volume generation experiment but with an additional fully connected RReLU layer of size $3,000$ in the encoder to compensate for the increased difficulty of the task. 

During training, images were given as input to the encoder which generated an activity vector of $200$ scalar values. These were divided in the \textit{shape code} comprising $185$ values and the \textit{transformation code} comprising $15$ values. The network was trained on $16,000$ image / volumes pairs with batches of size $10$. 

The switches connecting the encoder to the decoders were adjusted after every three training batches to allow the volume decoder and the image decoder to see the same number of examples. The volume decoder only received the \textit{shape code}, whereas the image decoder received both the \textit{shape code} and the \textit{transformation code}.

To test if the \textit{shape code} and the \textit{transformation code} learned the desired invariance we measured the mean standard deviation of activations for batches where only one of shape, pose or lighting conditions were changed. The same batches as in the invariance experiment were used. 

\begin{figure}[!t]
\hspace{-4mm}
  \begin{tabular}[b]{cc}
    \begin{subfigure}[b]{0.6\columnwidth}
      \includegraphics[width=\textwidth]{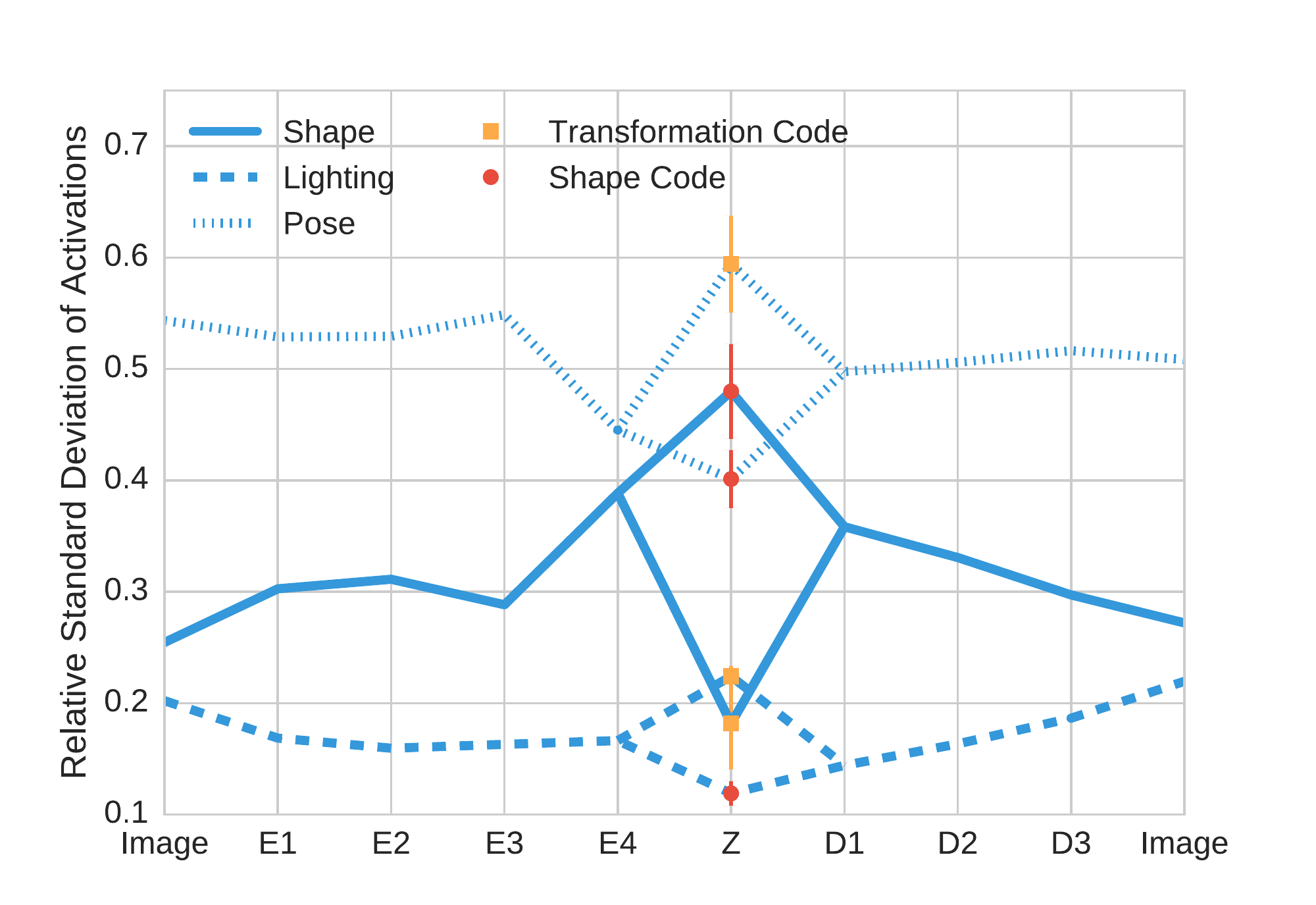}
 \vspace*{-8mm}
      \caption{}
      \label{fig:A}
    \end{subfigure}
\hspace{-7mm}
    \begin{tabular}[b]{c}
      \begin{subfigure}[b]{0.5\columnwidth}
        \includegraphics[width=\textwidth]{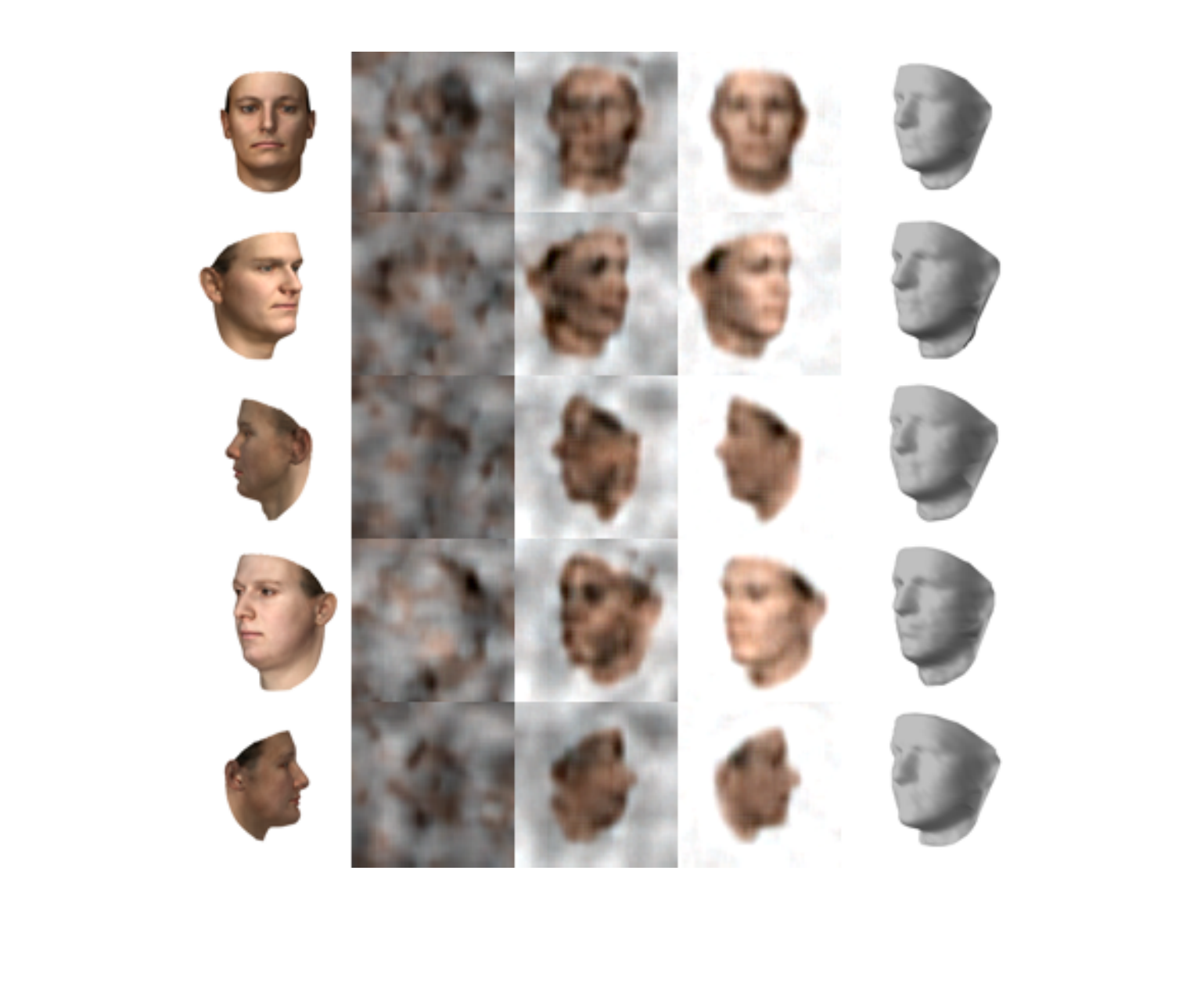}
        \vspace*{-10mm}
        \caption{}
        \label{fig:B}
      \end{subfigure}\\
    \end{tabular}
    \hspace{-10mm}
      \end{tabular}
    \hspace{-10mm}
\caption{\textbf{Disentangled representations:} (a) The relative mean standard deviation (SD) of activations in the encoder, \textit{shape code}, \textit{transformation code} and image decoder is compared for changes in shape, pose and lighting. The \textit{shape code} is most sensitive to changes in shape. The \textit{transformation code} is most sensitive to changes in pose and lighting. Error bars show standard deviation. (b) The output of the volume decoder and image decoder on a number of unseen images.  The first column is the input image. The second column is the image decoded from the \textit{shape code} only. The third column is the image decoded from the \textit{transformation code} only. The fourth column is the image decoded from the \textit{shape code} and the \textit{transformation code}. The fifth column is the output of the volume decoder shown from the same viewpoint for comparison.}
    \label{fig:disentangled}
\end{figure}

Figure~\ref{fig:disentangled}(a) shows the relative mean standard deviation of activations of each layer in the encoder, \textit{graphics code} and image decoder. The bifurcation at point Z on the plot shows that the two codes learned to respond differently to the same input.  The \textit{shape code} learned to be more sensitive to changes in shape than pose or lighting, and the \textit{transformation code} learned to be more sensitive to changes in pose and lighting than shape. 

To make sure the image decoder used the shape code to reconstruct the input we compared the output of the image decoder with input only from the \textit{shape code}, the \textit{transformation code} and both together. Figure~\ref{fig:disentangled}(b) shows the output of the volume decoder and image decoder on a number of unseen images.  The first column shows the input to the network. The second column shows the output of the image decoder with input only from the \textit{shape code}. The third column shows the same for the output of the \textit{transformation code}. The fourth column shows the combined output of the \textit{shape code} and the \textit{transformation code}. The fifth column shows the output of the volume decoder. 

\subsection{Face Recognition in Novel Pose and Lighting Conditions}
To measure the invariance and representational quality of the \textit{shape code} we tested it on a face recognition task. 

The point-wise Euclidean distance between the \textit{shape code} generated by an image was measured for a batch of $150$ random images including one image that was the same face with a different pose (target). The random images were ordered from the smallest to greatest distance and the rank of the target was recorded. This was repeated $100$ times and an identical experiment was performed for pose. The mean rank for the same face with a different pose was $11.08$. The mean rank of the same face with different lighting was $1.02$. This demonstrates that the \textit{shape code} can be used as a pose and lighting invariant face classifier.

To test if the \textit{shape code} was more invariant to pose and lighting than the full \textit{graphics code} we repeated this experiment using the full \textit{graphics code}. The mean rank for the same face with a different pose was $26.86$. The mean rank of the same face with different lighting was $1.14$. This shows that the \textit{shape code} was relatively more invariant to pose and lighting than the full \textit{graphics code}.

\subsection{Volume Predictions from Videos of Faces}
To test if video input improved the quality of the generated volumes we adapted the encoder to take video as input and compared to a single image baseline. $10,000$ video / volume pairs of faces were created. Each video consisted of five RGB frames of a face rotating from left facing profile to right facing profile in equidistant degrees of rotation. The same network architecture was used as in experiment 4.5. For the video model the first layer was adapted to take the whole video as input. For the single image baseline model, single images from each video were used as input.  

To test the performance difference between video and single image inputs a test set of 500 video / volume pairs was generated. Error was measured using the mean voxel-wise distance between ground truth and volumes generated by the network. For the video network the entire video was used as input. For the single image baseline each frame of the video was given separately as input to the network and the generated volume with the lowest error was used as the benchmark. 

A paired-samples t-test was conducted to compare error score in volumes generated from volumes and single images. There was a significant difference in the error score for video based volume predictions ($M=0.0073$, $SD=0.0009$) and single image based predictions ($M=0.0089$, $SD=0.0014$) conditions; $t(199)=-13.7522$, $1.0947e-30$.  

These results show that video input results in superior volume reconstruction performance compared with single images. 

\subsection{Volume Predictions from Images of Chairs}
In this experiment we tested the capacity of the network to generate volume predictions from objects with more variable geometry. $5000$ Volume / image pairs of chairs were created from the ModelNet dataset~\cite{wu20153d}. The images were $80 \times 80$ RGB images and the volumes were $30 \times 30 \times 30$ binary volumes. The predicted volumes were binarized with a threshold of $0.2$. Both decoders were used in this experiment. The \textit{shape code} consisted of $599$ activations and the \textit{transformation code} consisted of one activation. The \textit{shape code} was used to reconstruct the volumes. Both the \textit{shape code} and \textit{transformation code} were used to reconstruct the input.

\begin{figure}[!t]
  \hspace{-13mm}
  \begin{tabular}[b]{cc}
            \begin{subfigure}[b]{1.2\columnwidth}
      \includegraphics[width=\textwidth]{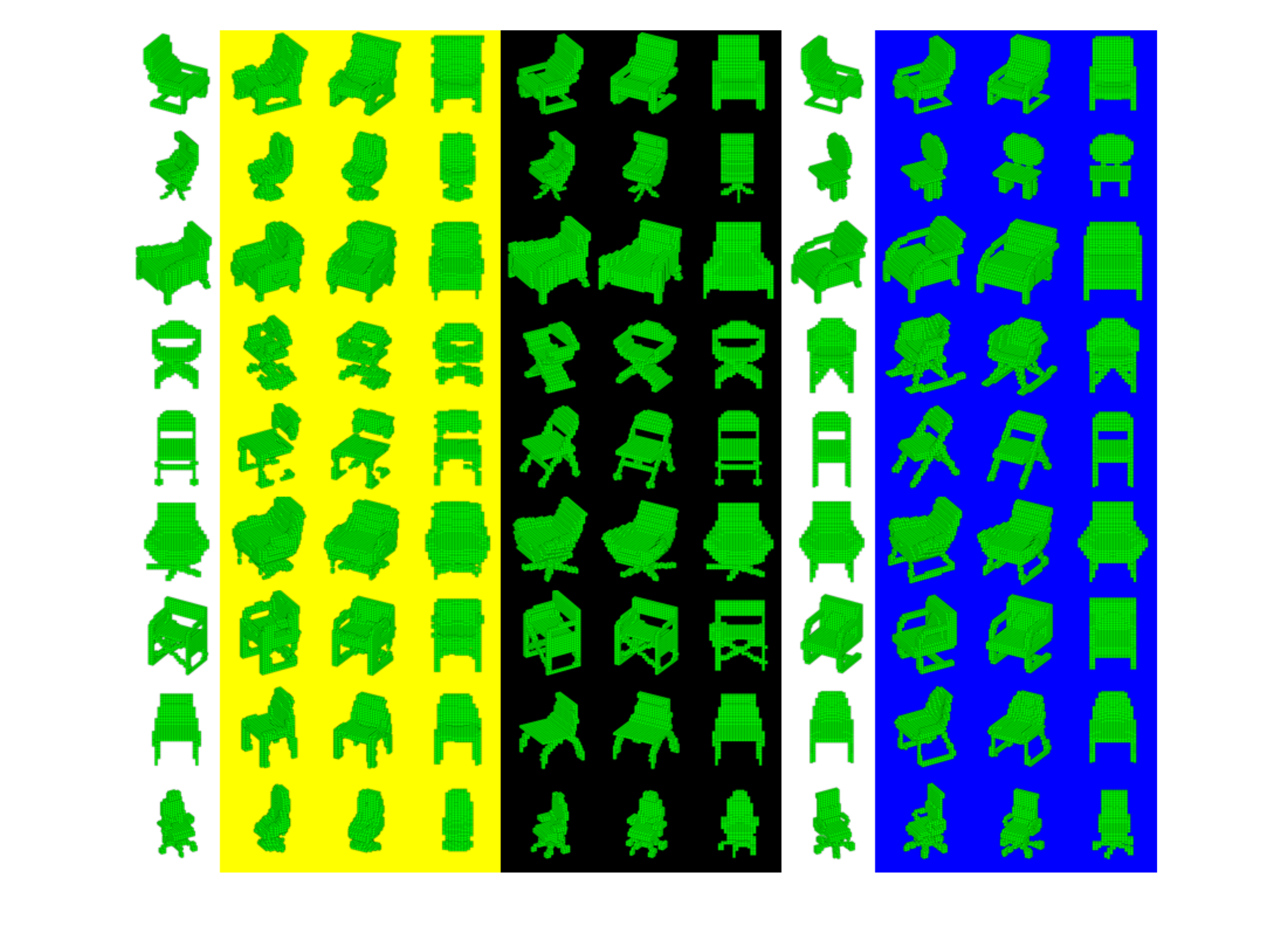}
      \vspace*{-10mm}
    \end{subfigure}\\
      \end{tabular}
  \caption{\textbf{Generated chair volumes:} Qualitative results showing the volume predicting capacity of the network on unseen data. First column: network inputs. Columns 2-4 (Yellow): network predictions shown from three viewpoints. Columns 5-7 (black): ground truth from the same viewpoints. Column 8: nearest neighbour image in the training set. Columns 9-11 (blue): nearest neighbour image ground truth.} 
   \label{fig:AAAA}
\end{figure}

Figure~\ref{fig:AAAA} demonstrates the network's capacity to generate volumetric predictions of chairs from novel images.

\subsection{Interpolating the Graphics Code}
\begin{figure}[!t]
  \hspace{-13mm}
  \begin{tabular}[b]{cc}
            \begin{subfigure}[b]{1.2\columnwidth}
      \includegraphics[width=\textwidth]{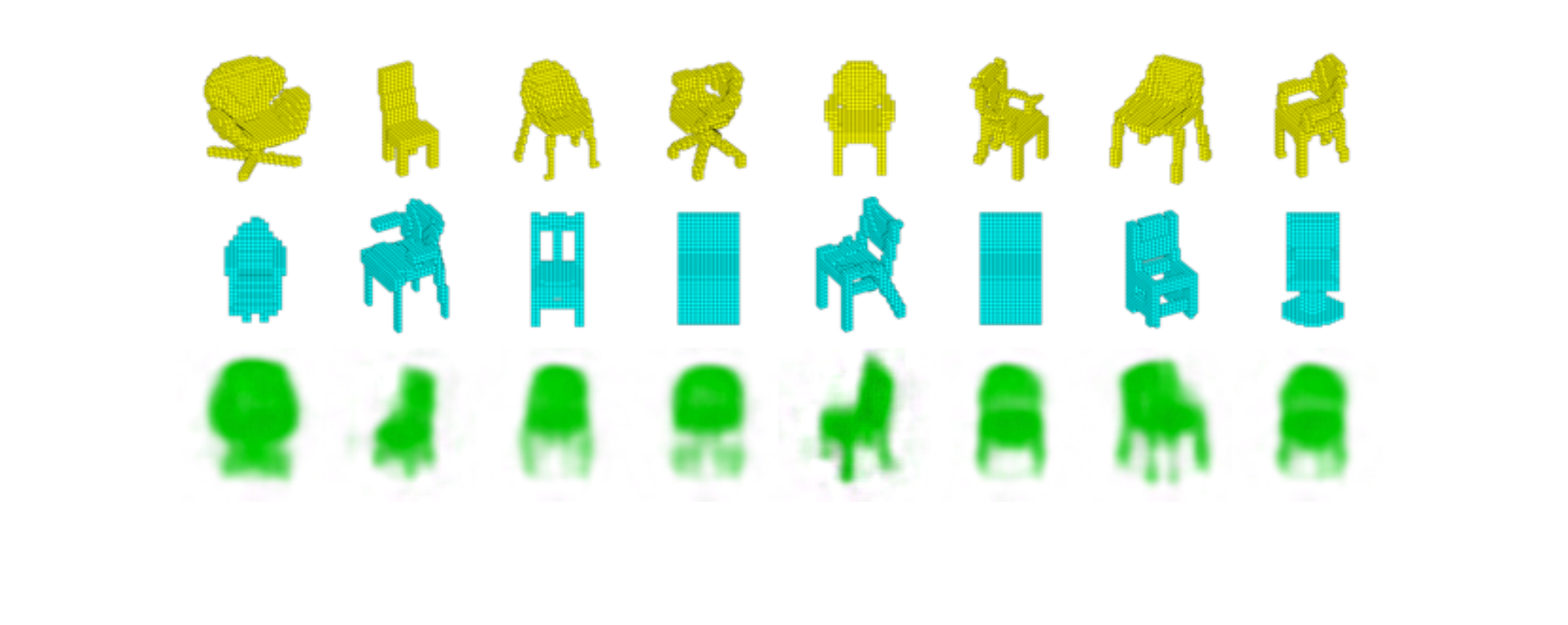}
      \vspace*{-10mm}
    \end{subfigure}\\
      \end{tabular}
  \caption{\textbf{Interpolated code:} Qualitative results combining the \text{shape code} and \textit{transformation code} from different images. First row: images used to generate the \textit{shape code}. Second row: images used to generate the \textit{transformation code}. Last row: Image decoder output.} 
   \label{fig:AAAAA}
\end{figure}

In order to qualitatively demonstrate that the \textit{graphics code} in experiment $4.8$ was disentangled with respect to shape and pose, we swapped the \textit{shape code} and \textit{transformation code} of a number of images and generated new images from the interpolated code using the image decoder. Figure~\ref{fig:AAAAA} shows the output of the image decoder using the interpolated code. The shape of the chairs in the generated images is most similar to the shape of the chairs in the images used to generate the \textit{shape code}. The pose of each chair is most similar to the pose of the chairs in the images used to generate the \textit{transformation code}. This demonstrates that the \textit{graphics code} is disentangled with respect to shape and pose.

\section{Discussion}
We have shown that a convolutional neural network can learn to generate a compact graphical representation that is disentangled with respect to shape, and transformations such as lighting and pose. This representation can be used to generate a full volumetric prediction of the contents of the input image.

By comparing the activations of batches corresponding with a specific transformation or the shape of the image, we showed that the network can learn to represent a \textit{shape code} that is relatively invariant to pose and lighting conditions. By adding an additional decoder to the network that reconstructs the input image, the network can learn to represent a \textit{transformation code} that represents the pose and lighting conditions of the input.

Extending the approach to real world scenes requires consideration of the viewpoint of the generated volume. Although the volume is invariant in the sense that it contains all the information necessary to render the generated object from any viewpoint, a canonical viewpoint was used for all volumes so that they were generated from a frontal perspective. Natural scenes do not always have a canonical viewpoint for reference. One possible solution is to generate a volume from the same viewpoint as the input. Experiments show that this approach is promising but further work is needed. 

In order to learn, the network requires image-volume pairs. This limits the type of data that can be used as volumetric datasets of sufficient size, or models that generate them are limited in number. A promising avenue for future work is incorporating a professional quality renderer into the decoder structure. This theoretically allows for 3D graphical representations to be learned, provided that the rendering process is approximately differentiable.
\newline
\\
\noindent
\textbf{Acknowledgements:} Thanks to Thomas Vetter for access to the Basel Face Model.

\clearpage

\bibliographystyle{splncs}
\bibliography{egbib}
\end{document}